# Tackling the Kidnapped Robot Problem via Sparse Feasible Hypothesis Sampling and Reliable Batched Multi-Stage Inference

Muhua Zhang, Lei Ma, *Senior Member, IEEE*, Ying Wu, Kai Shen, Deqing Huang, *Senior Member, IEEE*, and Henry Leung, *Fellow, IEEE*

***Abstract*—This paper addresses the Kidnapped Robot Problem (KRP), a core localization challenge of relocalizing a robot in a known map without a prior pose estimate upon localization loss or at SLAM initialization. For this purpose, a passive 2-D global relocalization framework is proposed. It estimates the global pose efficiently and reliably from a single LiDAR scan and an occupancy grid map while the robot remains stationary, thereby enhancing the long-term autonomy of mobile robots. The proposed framework casts global relocalization as a non-convex problem and solves it using a multi-hypothesis scheme with batched multi-stage inference and early termination, balancing completeness and efficiency. The traversability-constrained Rapidly-exploring Random Tree (RRT) asymptotically covers the reachable space and restricts the search space to traversable regions in the occupancy grid, generating sparse and uniformly distributed feasible positional hypotheses. The hypotheses are preliminarily ordered by the proposed Scan Mean Absolute Difference (SMAD), a coarse beam-error-level metric that enables efficient early termination by prioritizing high-likelihood candidates and is optimized for limited scan measurements. The Translation-Affinity Scan-to-Map Alignment Metric (TAM) is introduced for reliable orientation selection and accurate final pose evaluation, mitigating the degradation of conventional likelihood-field-based metrics under translational uncertainty, non-panoramic LiDAR scans, and environmental changes. Real-world experiments on a resource-constrained mobile robot with non-panoramic LiDAR scans in two representative indoor scenarios show that the proposed framework improves the mean relocalization success rate by about $26.9$ percentage points over the strongest baseline, while reducing runtime from several seconds or even tens of seconds to the sub-second to $1.6s$ range in most comparable cases. These results demonstrate that the proposed method achieves higher success rates with substantially lower runtime under the same time constraint, validating its effectiveness as a practical relocalization module for laser SLAM.**

***Index Terms*—Global relocalization, pose estimation, multiple hypotheses, non-panoramic LiDAR.**

## I. INTRODUCTION

WITH the advancement of measurement and robotics technology, automated measurement systems [1], [2], [3] and mobile robots [4], [5] are increasingly applied in industrial inspection and measurement tasks. Safe and efficient operation of mobile robots requires robust autonomous mapping, localization, and navigation [6], [7], [8], [9]. General localization methods include inertial odometry [10], vision-based Simultaneous Localization and Mapping (SLAM) [11], LiDAR-based SLAM, either standalone [12] or vision-augmented [13], [14], and infrastructure-based methods [15]. Overall, compared with vision, LiDAR-based SLAM is robust and has been widely adopted in practice [16], particularly in 2-D indoor localization tasks.

In LiDAR-based SLAM, the entire pure localization process usually happens at two stages [17]. The first is global relocalization, which provides a coarse pose estimate within the prior map and essentially corresponds to the Kidnapped Robot Problem (KRP), while the second is local refinement, enabling continuous pose tracking. Many studies aim to unify these stages. 2-D Monte Carlo Localization (MCL)-based methods [18], [19], [20], [21], [22] sample global pose hypotheses and update particle weights iteratively, expecting convergence to the correct pose. Scan-correlation-based systems such as Cartographer [23] and pose registration approaches with global matching capabilities like [24] typically search candidate poses within a bounded window and apply branch-and-bound (BnB) optimization. However, when the robot remains stationary, the likelihood landscape becomes highly non-convex. The periodic alignment between the LiDAR scan and map yields multiple observation-consistent local maxima. Thus, this multimodality breaks monotonicity and prevents convergence, violating both large-sample assumptions in particle filters and the unimodality required by BnB. As a result, these methods often fail or are slow to perform global relocalization when the robot is stationary. Therefore, in practice, global relocalization is often operated manually, following the paradigm of the Robot Operating System (ROS) [25], or assisted by external beacons. Consequently, when starting from an unknown pose or recovering from the localization failure, LiDAR-only mobile robots struggle to autonomously restore pose tracking, limiting long-term autonomy under minimal sensor configurations.

Many studies focus on global relocalization instead of unifying it with pose tracking in SLAM to address this challenge. Conventional methods can be categorized as active or passive. Active relocalization methods exploit robot motion to increase observation coverage for improved matching [26]

This work was supported by the National Natural Science Foundation of China (Grants 62203371). *(Corresponding author: Kai Shen).*

Muhua Zhang, Lei Ma, Ying Wu, Kai Shen, Deqing Huang are with the School of Electrical Engineering, Southwest Jiaotong University, Chengdu 611756, China (e-mail: shenkai@swjtu.edu.cn).

Henry Leung is with the Department of Electrical and Computer Engineering, University of Calgary, Calgary AB T2N 1N4, Canada (e-mail: leungh@ucalgary.ca).

or to reduce pose uncertainty [27], [28], [29], [30], [31]. They improve the chance of converging to the correct global pose but require robot motion, reducing efficiency and increasing risk in unknown environments. Passive relocalization estimates the global pose directly from observations without significant degradation while the robot remains stationary. It is more efficient and safer when the environment is generally discriminative, making it an emerging research direction. A LiDAR scan descriptor for passive global relocalization is proposed in [32]. It is effective in 3-D point cloud maps but dependent on laser intensities, which are unstable over long-term spatiotemporal evolution and unsuitable for common 2-D indoor SLAM. The 2-D method [33] compared LiDAR scan images with multi-scale map images using line and point features for pre-filtering and further leveraged a known relation between the Attitude and Heading Reference System (AHRS) heading and the map orientation to reduce the search space. While this strategy can effectively narrow down candidate poses under favorable conditions, image feature-based matching may be sensitive to scene complexity in real-world settings. In addition, the orientation pruning relies on the availability of reliable AHRS alignment, which may not always be accessible in practical industrial deployment scenarios. Recent works sample candidate poses across the map to form a search space, then apply staged scoring and selection to progressively narrow it. The main challenges include balancing sample density and map coverage, improving robustness of scan-to-map alignment metrics, and maintaining efficiency and solution diversity in candidate selection. The related work [34] uses particle-based random sampling to generate hypothesis candidates and performs brute force search scoring on each candidate through Fourier-Mellin invariant matching. However, traditional particle sampling struggles to balance uniform coverage with sample size, and ignores traversability, producing several invalid hypotheses. The efficiency of the approach is limited when it is combined with brute force search. The work in [35] adopts a similar sampling strategy but introduces the Cumulative Absolute Error per Ray (CAER) to measure scan-to-map alignment. It uses CAER for pre-filtering and re-evaluates refined poses after registration. This improves efficiency, but for non-panoramic LiDAR scan, CAER lacks rotational invariance, requiring enumeration of orientations and beam-wise recomputation. However, staged selection based merely on coarse CAER may still miss the correct pose.

Motivated by the above challenges, this paper studies passive global relocalization for a stationary mobile robot using only a single 2-D LiDAR scan and a prior grid map when localization fails or no reliable prior pose is available. This relocalization is performed before continuous pose tracking resumes. To this end, a 2-D passive global relocalization framework is proposed. The framework is generally applicable to 2-D LiDAR-based passive global relocalization under sufficiently observable conditions and remains effective even for resource-constrained robots equipped with non-panoramic LiDAR. In cases where the environment is inherently ambiguous, the relocalization problem itself becomes ill-posed for any single-scan passive method. The framework serves as a practical recovery or initialization module in laser SLAM and consists of four key components, namely feasible positional hypothesis generation, fast coarse hypothesis ordering, reliable fine pose evaluation, and batched multi-stage inference. The main contributions of this paper are summarized as follows:

1) A traversability-constrained Rapidly-exploring Random Tree (RRT) sampling strategy is introduced to restrict the search space to the robot's reachable region and to generate sparse yet spatially distributed positional hypotheses. This component ensures that all sampled hypotheses are feasible and provides high-quality candidates for subsequent relocalization inference.
2) A coarse beam-error-level metric named Scan Mean Absolute Difference (SMAD) is proposed for rapid scan-to-map similarity evaluation and preliminary hypothesis ordering. For non-panoramic LiDAR scans, SMAD further exploits prefix-sum-based acceleration of map-synthesized scan ranges to handle orientation-wise recomputations caused by angular anisotropy, thereby supporting efficient early termination.
3) An enhanced likelihood-field-based metric named Translation-affinity Scan-to-Map Alignment Metric (TAM) is proposed for robust orientation selection and accurate final pose evaluation under sparse sampling, non-panoramic LiDAR scans, and scene dynamics. By integrating an FoV-adaptive beam retention mechanism and a structural consistency term, TAM improves the reliability of likelihood-field-based evaluation under challenging relocalization conditions.
4) A batched multi-stage inference process with local $top-k$ selection and early termination is proposed to progressively filter and verify pose candidates in the non-convex global relocalization problem. This design balances diversity, reliability, and computational efficiency, making the proposed framework practical for real-world deployment.

Real-world experiments and ablation studies demonstrate that the proposed framework improves relocalization success rate and runtime efficiency compared with representative baselines, validating its effectiveness as a practical module for laser SLAM that supports initialization without a prior pose estimate and recovery after localization failure.

The remainder of this paper is organized as follows. Section II formulates passive global relocalization. Section III presents the proposed relocalization framework. Section IV reports real-world experiments and ablation studies. Section V gives the conclusion of the paper and potential future work.

## II. Preliminaries and Problem Statement

### A. Passive Global Relocalization

In the passive global relocalization, the robot remains stationary and infers its global pose exclusively from a LiDAR scan and an occupancy-grid map without an initial pose estimate [33], [34], [35]. Let the map be denoted by $\boldsymbol{\mathcal{M}}$, and the free space in $\boldsymbol{\mathcal{M}}$ by $\Omega \subset \mathbb{R}^2$. The robot pose is defined as

$$\boldsymbol{x} = (\boldsymbol{p}, \theta) \in \mathbb{SE}(2), \boldsymbol{p} = (x, y)^T \in \Omega, \theta \in [-\pi, \pi). \quad (1)$$

A single LiDAR scan is represented as

$$\boldsymbol{\mathcal{Z}} = (\phi_i, z_i)_{i=1}^{N_Z}, \phi_i \in [-\pi, \pi), z_i \in \mathbb{R}^+, \quad (2)$$

where $N_Z$ is the number of beams, $\phi_i$ the beam angle, and $z_i$ the corresponding range on $\phi_i$.

The global relocalization process can be formulated as a maximization of a generalized evaluation function [33]:

$$\boldsymbol{x}^* = \underset{x}{argmax}\, \mathcal{F}(\boldsymbol{\mathcal{Z}}, \boldsymbol{x} \mid \boldsymbol{\mathcal{M}}), \quad (3)$$

where $\mathcal{F}$ outputs an alignment score between the LiDAR scan observation $\boldsymbol{\mathcal{Z}}$ and the map $\boldsymbol{\mathcal{M}}$ under a candidate pose $\boldsymbol{x}$. $\mathcal{F}$ is generally designed to be implemented based on likelihood field [33] or based on beam-error level metric [35]. This problem exhibits non-convexity and multimodality. Periodic alignments between environmental structures and scan beams lead to multiple local maxima, especially under non-panoramic LiDAR FoV. Formally, there may exist a set of stationary points $\{\boldsymbol{x}^{(k)}\}_{k=1}^{K}$ such that

$$\nabla_x \mathcal{F}\big(\boldsymbol{\mathcal{Z}}, \boldsymbol{x}^{(k)} \mid \boldsymbol{\mathcal{M}}\big) = 0, \nabla_x^2 \mathcal{F}\big(\boldsymbol{\mathcal{Z}}, \boldsymbol{x}^{(k)} \mid \boldsymbol{\mathcal{M}}\big) \prec 0, \quad (4)$$

with

$$\mathcal{F}\big(\boldsymbol{\mathcal{Z}}, \boldsymbol{x}^{(1)} \mid \boldsymbol{\mathcal{M}}\big) \approx \mathcal{F}\big(\boldsymbol{\mathcal{Z}}, \boldsymbol{x}^{(2)} \mid \boldsymbol{\mathcal{M}}\big) \approx \ldots \approx \mathcal{F}\big(\boldsymbol{\mathcal{Z}}, \boldsymbol{x}^{(K)} \mid \boldsymbol{\mathcal{M}}\big). \quad (5)$$

Consequently, this multimodal landscape violates the unimodality assumption required for effective convergence, often causing gradient-based optimization solutions to stall in local optima. This makes it difficult for global relocalization to directly use Iterative Closest Point (ICP) or Normal Distributions Transform (NDT) without initial pose guesses.

*Remark 1*: The formulation (3) relies on the premise that the evaluation function yields a dominant and distinguishable optimum corresponding to the true robot pose, which generally holds when the measurement contains sufficient structural information. However, this premise may degrade under limited observability when only a single LiDAR scan is available.

In environments with strong structural symmetries (e.g., long corridors or repetitive layouts), limited FoV coverage, or scans dominated by free space, different candidate poses may produce highly similar evaluation scores, making them difficult to distinguish from a single observation. In highly symmetric or largely open corridor environments, the proposed framework and passive baselines may also fail to produce a unique and reliable solution. In addition, significant environmental changes may reduce the consistency between the scan and the map. These cases indicate that the ambiguity arises from the measurement condition itself rather than a specific algorithm. When such ambiguity occurs, additional information, typically through robot motion, is required to resolve competing hypotheses, as in active relocalization frameworks such as multiple hypothesis tracking [28].

Nevertheless, real-world challenging environments are typically not perfectly degenerate and still exhibit subtle structural variations. As demonstrated in the experimental section, the proposed framework can exploit such subtle differences, enabling effective disambiguation in scenarios with door state changes as well as in corridor environments that are visually similar but not strictly symmetric. Under such conditions, passive relocalization provides a fast and safe initialization and can further serve as a source of high-quality candidate poses for subsequent multi-hypothesis tracking.

### *B. Multiple Hypotheses-Based Global Relocalization*

Many approaches use discretized form to solve the global pose estimation problem. Based on discrete pose hypotheses, [18], [19], [20], [21], [22] use particle filter-based MCL to converge to the global pose. Based on discrete pose search windows, [23] and [24] use BnB-accelerated scan matching to achieve pose registration. Particle filter-based MCL struggles in the stationary case because particles cannot diffuse via the motion model and rely only on observation likelihood. In multimodal landscapes, this causes convergence to local optima. Moreover, BnB-based methods, on the other hand, may rule out the correct solution direction early on, or turn into the brute-force search due to matching fitness score degradation.

Recent works on global relocalization like [34], [35] combine randomly sampled position hypotheses with a set of angles enumerated at the positions to form pose hypotheses. A set of sampled positions with the size of $N_p$ is generated as

$$\boldsymbol{\mathcal{P}} = \{\boldsymbol{p}_j\}_{j=1}^{N_p}, \boldsymbol{p}_j \in \Omega, \quad (6)$$

For each position $\boldsymbol{p}_j$, the orientation is determined by solving:

$$\theta_j^* = \underset{\theta \in [-\pi, \pi)}{argmax}\, \mathcal{F}\big(\boldsymbol{\mathcal{Z}}, (\boldsymbol{p}_j, \theta) \mid \boldsymbol{\mathcal{M}}\big), \quad (7)$$

where the optimization can be performed either through discretized enumeration of an orientation set or by continuous optimization over the orientation domain. The resulting pose hypothesis set $\mathcal{H}$ can be expressed as

$$\mathcal{H} = \{(\boldsymbol{p}_j, \theta_j^*)\}. \quad (8)$$

The global relocalization problem (3) is reformulated as selecting the best hypothesis from $\mathcal{H}$:

$$(\boldsymbol{p}, \theta)^* = \underset{(\boldsymbol{p}_j, \theta_j^*) \in \mathcal{H}}{argmax}\, \mathcal{F}\big(\boldsymbol{\mathcal{Z}}, (\boldsymbol{p}_j, \theta_j^*) \mid \boldsymbol{\mathcal{M}}\big). \quad (9)$$

A common way to construct $\boldsymbol{\mathcal{P}}$ is through fixed-amount random particle sampling. But this faces several drawbacks:

1) When $N_p$ is increased to improve sampling coverage and uniformity, the search space increases, which leads to a fundamental increase in computational complexity.
2) Sampled particles may lie within the free space Ω, but not within the robot's traversable set $\mathcal{R}(\boldsymbol{p}_0)$, i.e., there exists no feasible trajectory from the map origin $\boldsymbol{p}_0$ to the sampled position under collision constraints. Such unreachable hypotheses introduce invalid evaluations and unnecessary computational overhead.

Therefore, uniform and sparse sampling which ensures that the sampling positions are all within $\mathcal{R}(\boldsymbol{p}_0)$ can fundamentally improve computational efficiency. However, sparse sampling inevitably introduces translational error between the correct position $\boldsymbol{p}^*$ and the nearest sampled position $\boldsymbol{p}_j$:

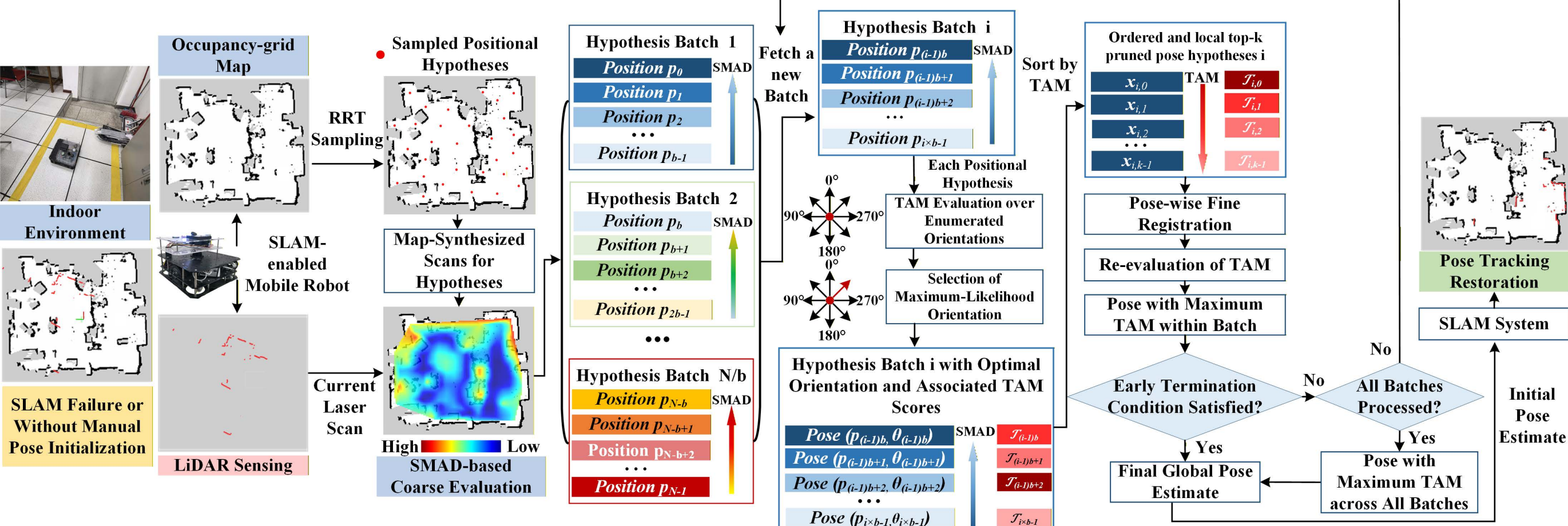

**Fig. 1.** The entire pipeline of the proposed framework.

---

**Algorithm 1** Traversability-Constrained Hypothesis Sampling

---

1: **Input:** Map origin $\boldsymbol{p}_0$ , RRT tree $\boldsymbol{T}_{RRT}$, RRT sampling boundary $\boldsymbol{\mathcal{B}}$, grid map $\boldsymbol{\mathcal{M}}$, map width $W_{\mathcal{M}}$, map height $H_{\mathcal{M}}$, map resolution $r_{\mathcal{M}}$, sampling spacing $\rho$, and gain threshold $\varepsilon_{gain}$

2: **Output:** Sparse hypotheses $\boldsymbol{\mathcal{P}} = \{\boldsymbol{p}_j\}_{j=1}^{N_p}$ with $\boldsymbol{p}_j \in \mathcal{R}(\boldsymbol{p}_0)$

3: **Initialize:** $\boldsymbol{T}_{RRT} \leftarrow \{\boldsymbol{p}_0\}$, $\boldsymbol{\mathcal{P}} \leftarrow \emptyset$, $\boldsymbol{\mathcal{P}}_{last} \leftarrow \emptyset$

4: $N_{sampling} \leftarrow \lceil W_{\mathcal{M}} H_{\mathcal{M}} {r_{\mathcal{M}}}^2 / \rho^2 \rceil$

5: **do**

6:   **for** $\boldsymbol{i} \leftarrow 1$ to $N_{sampling}$ **do**

7:     $\boldsymbol{p}_{rand} \leftarrow Random\ point$

8:     $\boldsymbol{p}_{nearest} \leftarrow NearestNode(\boldsymbol{p}_{rand}, \boldsymbol{T}_{RRT})$

9:     $\eta \leftarrow AdaptiveExpandDist(\boldsymbol{p}_{nearest}, \boldsymbol{\mathcal{M}})$

10:     $\boldsymbol{p}_{new} \leftarrow ExtendPoint(\boldsymbol{p}_{rand}, \boldsymbol{p}_{nearest}, \eta)$

11:     **if not** ($InSamplingBoundary(\boldsymbol{p}_{new}, \boldsymbol{\mathcal{B}})$ **and** $TraversabilityCheck(\boldsymbol{p}_{nearest}, \boldsymbol{p}_{new}, \boldsymbol{\mathcal{M}})$) **then**

12:       **continue**

13:     $\boldsymbol{T}_{RRT} \leftarrow \boldsymbol{T}_{RRT} \cup \{\boldsymbol{p}_{new}\}$

14:   $\boldsymbol{\mathcal{P}}_{last} \leftarrow \boldsymbol{\mathcal{P}}$

15:   $\boldsymbol{\mathcal{P}} \leftarrow VoxelDownsample(\boldsymbol{T}_{RRT}, \rho)$

16: **while** $|\boldsymbol{\mathcal{P}} \setminus \boldsymbol{\mathcal{P}}_{last}| > \varepsilon_{gain}$

17: **return** $\boldsymbol{\mathcal{P}}$

---

$$\Delta \boldsymbol{p}_j = \boldsymbol{p}^* - \boldsymbol{p}_j, \|\Delta \boldsymbol{p}_j\| \leq \rho, \tag{10}$$

where $\rho$ is determined by the sampling resolution. $\mathcal{F}$ must therefore exhibit translational affinity, i.e., it should retain orientation discriminability and pose separability even under bounded translational deviations. At the same time, in the real world, the environment often changes, and there are differences between the map and the observation. If the overall output of $\mathcal{F}$ at the correct pose $\boldsymbol{x}^*$ degenerates to a small value, the unified threshold benchmark will become invalid, and the correct solution will be indistinguishable from the incorrect solution. Therefore, $\mathcal{F}$ also needs to have dynamic robustness.

For the full pipeline, the global $top - k$ pruning-selection framework used in [34], [35] scores all candidate poses and retains only the $top - k$ hypotheses for the next round. The process iterates through several rounds of ranking and selection before producing the final solution. However, the drawback lies in its limited inclusiveness.

1) Due to the uncertainty of the evaluation metric caused by $\Delta \boldsymbol{p}_j$, hypotheses near the correct position may rank high only among the samples with similar metric values (i.e., local $top - k$) rather than in the global $top - k$. If global pruning is performed, the correct solution may be mistakenly eliminated at the early stage.

2) The process inherently constructs a biased search that depends more on early ranking results than on the global structure of $\mathcal{F}$, which may lead to convergence toward incorrect peaks.

In summary, challenges in sparse and uniform sampling, robustness of evaluation metrics, and inclusiveness of inference pipelines remain in existing multi-hypothesis-based methods, providing the motivation for this work.

## III. Proposed Framework

Fig. 1 illustrates the overall process of the proposed 2-D passive global relocalization framework. The robot starts from an unknown non-origin pose without manual initialization. The framework only takes a single LiDAR scan and an occupancy grid map as inputs. The traversability-constrained RRT first performs sparse and uniform sampling in the reachable free space to obtain global candidate positions. The beam-error level coarse metric SMAD is then used to compute the similarity between the LiDAR scan and a map-synthesized virtual scan (obtained by ray-casting the map at the positional hypothesis) for each position, rank them globally, and divide the ordered set into batches. For each batch, candidate positions are combined with enumerated orientations and evaluated using the likelihood-field-based precision alignment metric TAM. The orientation with the highest score is selected for each position, resulting in a TAM-ordered set of pose hypotheses. $Top - k$ poses are further aligned using ICP between the LiDAR scan and the entire map, followed by re-evaluations with TAM. The system terminates early if the best re-evaluated TAM score $\hat{\mathcal{T}}_b^*$ exceeds a threshold $\tau$. Otherwise, batches are processed sequentially, and the final pose is taken as the highest-scoring hypothesis across all batches, which is then fed back to the SLAM system to resume continuous pose tracking.

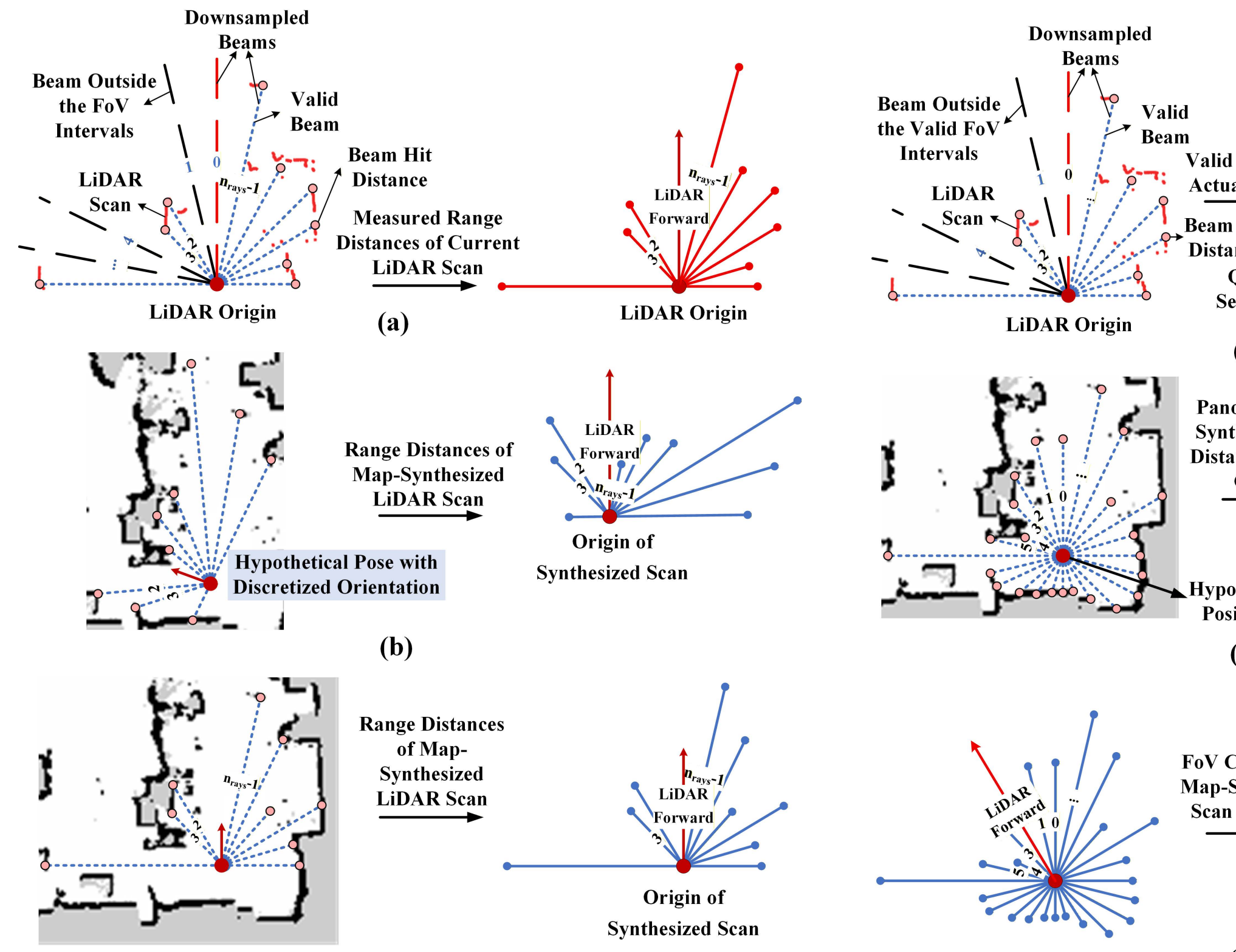

**Fig. 2.** Anisotropic orientation characteristic of beam-error level metrics.

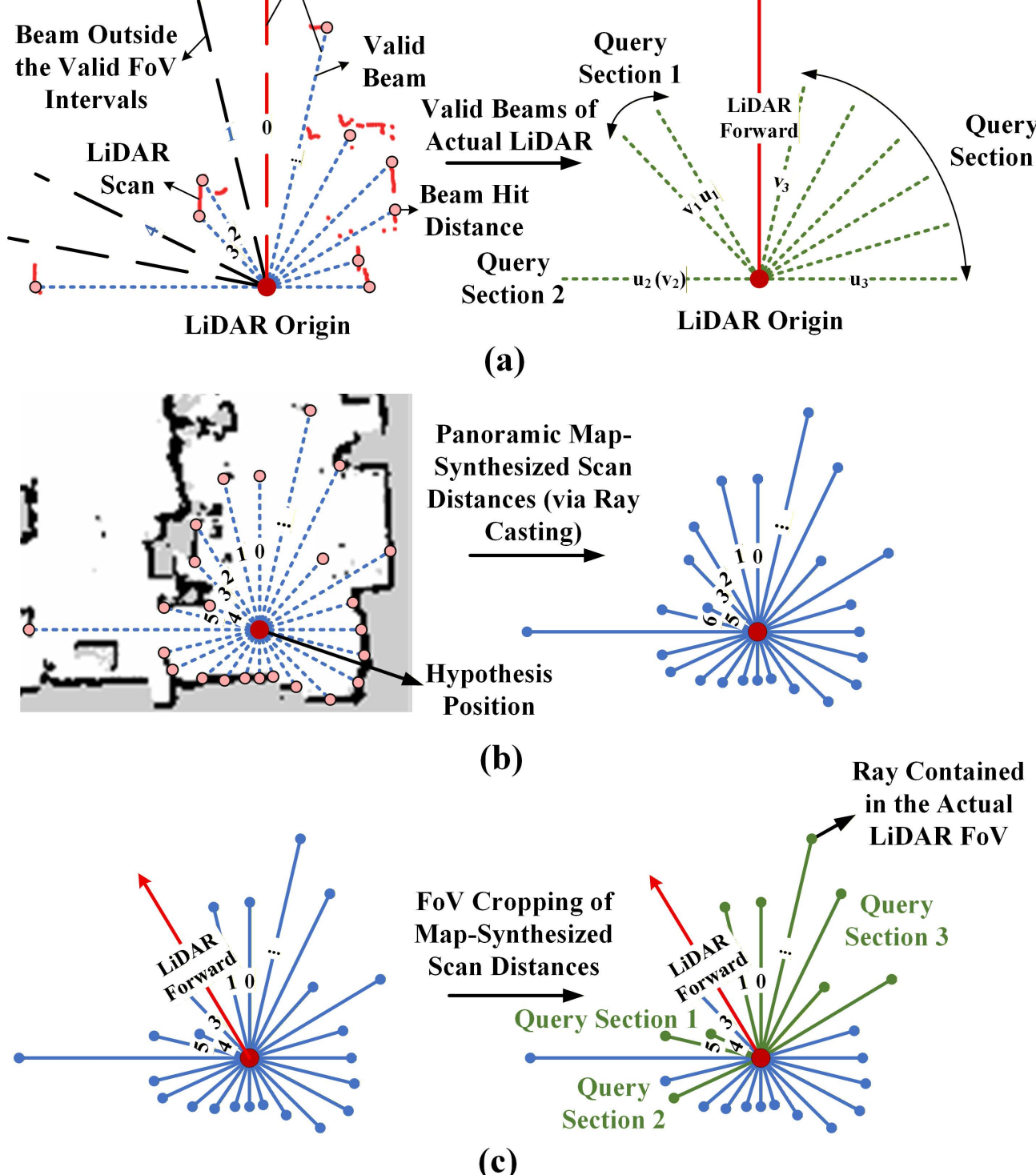

**Fig. 3.** Accelerated SMAD calculation via prefix sum of ranges from the panoramic map-synthesized scan.

*Remark 2*: In relocalization methods using pure $top-k$ selection [34], [35], let the correct pose be $\boldsymbol{x}^* = (\boldsymbol{p}^*, \theta^*)$, and define $\boldsymbol{\mathcal{B}}^* = \{\boldsymbol{x}_j : \mathcal{G}^t(\boldsymbol{x}_j) \to \boldsymbol{x}^*\}$ as the set of hypotheses that converge to $\boldsymbol{x}^*$ under a local refinement operator $\mathcal{G}$ (e.g., ICP). For any $\boldsymbol{x} \in \boldsymbol{\mathcal{B}}^*$, let $r(\boldsymbol{x})$ denote its rank index under the discrete scoring $\mathcal{F}\big(\boldsymbol{\mathcal{Z}}, (\boldsymbol{p}_j, \theta_j^*)\,\big|\,\boldsymbol{\mathcal{M}}\big)$ evaluated at discrete positions $\boldsymbol{p}_j$ with the maximum-likelihood orientation $\theta_j^*$. Since each $\boldsymbol{p}_j$ deviates from $\boldsymbol{p}^*$ by the non-uniform offset $\Delta\boldsymbol{p}_j$, we can write $\mathcal{F}\big(\boldsymbol{\mathcal{Z}}, (\boldsymbol{p}_j, \theta_j^*)\,\big|\,\boldsymbol{\mathcal{M}}\big) = \mathcal{F}(\boldsymbol{\mathcal{Z}}, (\boldsymbol{p}^*, \theta^*) \mid \boldsymbol{\mathcal{M}}) + \delta(\Delta\boldsymbol{p}_j) + \varepsilon_j$, where the bias $\delta(\Delta\boldsymbol{p}_j)$ and noise $\varepsilon_j$ vary across $j$. Hence, no evaluation metric on discrete hypotheses possesses strict global consistency for comparison, and the event that all convergent hypotheses are pruned, $\mathcal{E}_{drop}^{prune} = \{\min_{\boldsymbol{x} \in \boldsymbol{\mathcal{B}}^*} r(\boldsymbol{x}) > k\}$, occurs with a non-negligible probability. The proposed local $top-k$ batch inference reduces this risk by selecting the $top-k$ hypotheses from each score-level region, thereby ensuring that $\boldsymbol{x}^*$ and its neighboring hypotheses within the same level are more likely to be explicitly evaluated before any possible early termination. The early-termination threshold $\tau$ is initially motivated by well-localized score levels and validated by experiments.

*Remark 3*: Although the proposed pipeline does not perform aggressive search pruning, its efficiency is preserved by traversability-constrained uniform sparse RRT sampling that reduces the hypothesis domain, ranking based on SMAD that prioritizes promising candidates for early termination, and TAM that provides reliable orientation selection under sparse hypotheses. Compared with sequential pure global $top-k$ selection, the proposed pipeline is inherently parallelizable, leading to favorable time costs under efficient implementation.

### *A. Traversability-Constrained Hypothesis Sampling*

When the hypothesis sampling process needs to maintain both randomness and traversability, RRT is an effective choice. It samples random points in the feasible space and expands along traversable directions. This balances global exploration and local connectivity. The RRT's asymptotic completeness ensures probabilistic coverage of traversable space as iterations increase. It provides a sparse and uniformly distributed set of feasible hypotheses for global relocalization. The proposed framework uses an enhanced full-coverage RRT for positional hypothesis sampling. During node expansion, the expansion distance is adaptively adjusted based on the shortest distance between the node and nearby obstacles. The step size decreases linearly when approaching obstacles, improving exploration in narrow areas. A strict traversability check is applied. If any point along a candidate path is closer to an obstacle than the robot radius $r_{robot}$, the expansion is rejected to prevent unreachable samples. Batched expansion and gain-based termination on downsampled nodes create an efficient stopping mechanism. This design removes the need for time-based execution to ensure coverage completeness. All RRT nodes are uniformly downsampled to generate the final set of positional hypotheses. Algorithm 1 presents the detailed

process with:

1) *NearestNode:* Returns the node in $\boldsymbol{T}_{RRT}$ that minimizes the Euclidean distance to the input point $\boldsymbol{p}$.
2) *ExtendPoint:* Generates a new point from $\boldsymbol{p}_{nearest}$ toward $\boldsymbol{p}_{rand}$ with expansion distance $\eta$.
3) *MinDistToObstacle:* Computes the minimum Euclidean distance from the input point to the nearest occupied cell on $\boldsymbol{\mathcal{M}}$. The distance is obtained from a precomputed lookup table built using EDT algorithm.
4) *InSamplingBoundary:* Checks whether the input point is within the sampling boundary $\boldsymbol{\mathcal{B}}$, where $\boldsymbol{\mathcal{B}}$ denotes the bounding rectangle of $\boldsymbol{\mathcal{M}}$.
5) *AdaptiveExpandDist:* Computes the adaptive expansion distance $\eta$ from the input point $\boldsymbol{p}$ according to $d_{obs} = MinDistToObstacle(\boldsymbol{p}, \boldsymbol{\mathcal{M}})$. When $d_{obs} \geq 2r_{robot}$, $\eta = \eta_{max}$, the maximum expansion distance of RRT. When $d_{obs} \leq r_{robot}$, $\eta = r_{robot}$.When $r_{robot} \leq d_{obs} \leq 2r_{robot}$, $\eta$ decreases linearly as $d_{obs}$ decreases.
6) *TraversabilityCheck:* Verifies whether all points $\boldsymbol{p}$ on the line segment $[\boldsymbol{p}_0, \boldsymbol{p}_1]$ satisfy $d_{obs}(\boldsymbol{p}) \geq r_{robot}$.
7) *VoxelDownsample:* Performs voxel downsampling of $\boldsymbol{T}_{RRT}$ with sampling spacing $\rho$.

*Remark 4*: When the entire map $\boldsymbol{\mathcal{M}}$ is traversable, the number of representative points required for full coverage at sampling spacing $\rho$ is $N_{sampling} = \lceil W_{\mathcal{M}} H_{\mathcal{M}} r_{\mathcal{M}}{}^2/\rho^2 \rceil$. Utilizing $N_{sampling}$ expansion attempts per iteration matches the density of complete coverage, ensuring sufficient exploration intensity. The incremental coverage for the $t^{th}$ iteration is defined as $gain(t) = \left| \boldsymbol{\mathcal{P}}^{(t)} \setminus \boldsymbol{\mathcal{P}}^{(t-1)} \right|$. When $gain(t) \leq \varepsilon_{gain} \to 0$, the probability of newly sampled points hitting uncovered regions approaches zero, indicating saturation of coverage over $\mathcal{R}(\boldsymbol{p}_0)$. A small positive $\varepsilon_{gain}$ maintains completeness while avoiding stochastic termination instability. This mechanism achieves adaptive control of coverage across arbitrary map scales and sampling spacing, realizing asymptotic quasi-completeness of coverage without explicit dependence on sampling duration.

*B. Beam-Error Level Coarse Metric SMAD*

Beam-error metrics compute per-beam distance errors to assess consistency between LiDAR scans. For each pose hypothesis, a map-synthesized scan is generated by ray casting on the occupancy grid map. Comparing it with the actual scan yields a score that reflects the proximity of the hypothesis to the correct pose [35]. Unlike likelihood-field metrics, beam-error level metrics are computationally efficient, numerically stable, and directly correlated with range measurements, eliminating the need for normalization. They are suitable for fast evaluation over large hypothesis spaces. Although multiple high-response hypotheses may occur, their efficiency enables effective preliminary ordering, allowing hypotheses near the correct pose to be prioritized for early termination. However, for the non-panoramic LiDAR, beam-error level metrics exhibit orientation anisotropy. The FoV of the map-synthesized scan must match that of the real LiDAR. Panoramic LiDAR maintains rotational invariance since it can cover the full scene regardless of orientation. In contrast, as shown in Fig. 2(a), non-panoramic scan only observes a partial region of the panoramic scan, and this visible region changes with orientation, as shown in Fig. 2(b) and Fig. 2(c). Thus, even for the same position, the beam set varies with orientation, introducing anisotropy. For such LiDAR scan, beam subsets must be reselected, and the metric recomputed across discrete orientations, which increases computational cost.

To address the efficiency issue, the proposed SMAD is defined as the absolute value of the difference between the average ranges of two scans. This allows for acceleration by prefixing the sum of the ranges of beams in map-synthesized scan. SMAD uses downsampled scan, which is represented as

$$\boldsymbol{\mathcal{Z}}' = (\phi_i', z_i')_{i=1}^{N_{\mathcal{Z}}'}, \phi_i' = \phi_{i \cdot n_{skip}^{beam}}, \tag{11}$$

where $n_{skip}^{beam}$ is the scan sampling stride. For non-panoramic LiDAR scan, the valid beam set is defined by the intrinsic FoV, which can be modeled as a set of predefined angular intervals, with a total number of $K$:

$$\boldsymbol{\mathcal{Q}}_{FoV} = \{\boldsymbol{I}_k = [u_k, v_k]\}_{k=1}^{K}, \boldsymbol{I}_k \cap \boldsymbol{I}_{k'} = \emptyset, k \neq k'. \tag{12}$$

Each $\boldsymbol{I}_k$ represents a contiguous segment of valid beams, determined by the sensor configuration and static occlusions (e.g., robot body or mounting structure). In practice, these intervals remain fixed across scans and directly define the valid beam subsets used for evaluation, as shown in Fig. 3(a). Given a positional hypothesis $\boldsymbol{p} \in \mathcal{R}(\boldsymbol{p}_0)$, a panoramic map-synthesized scan is first generated via ray casting, as shown in Fig. 3(b):

$$\widehat{\boldsymbol{\mathcal{Z}}}'(\boldsymbol{p}) = \{\hat{z}_j'(\boldsymbol{p})\}_{j=1}^{N_s}, \tag{13}$$

where $N_s$ is the total number of beams. It can be determined by the LiDAR angular resolution $\Delta\phi$ and $n_{skip}^{beam}$:

$$N_s = \frac{2\pi}{\Delta\phi \cdot n_{skip}^{beam}}. \tag{14}$$

This ensures one-to-one index correspondence between real and synthesized beams. For non-panoramic LiDAR, multiple orientations are enumerated for each positional hypothesis, and the minimum SMAD value among them is taken as the final SMAD score. LiDAR forward orientations $\Theta$ are enumerated by shifting beam indices over the panoramic scan:

$$\Theta = \{\theta_m | \theta_m = m(\Delta\phi \cdot n_{skip}^{beam})\}, \tag{15}$$

where $\mathrm{m} = \{0, n_{skip}^{enum}, 2n_{skip}^{enum}, \dots, N_s - 1\}$, and $n_{skip}^{enum}$ is the orientation enumerating stride. Each LiDAR orientation $\theta_m$ corresponds to an integer offset $m$. And orientation rotation can be achieved by shifting of $m$.

The mean range of the actual scan is computed once:

$$\bar{z}' = \frac{1}{\sum_k |\boldsymbol{I}_k|} \sum_{k=1}^{K} \sum_{i \in \boldsymbol{I}_k} z_i'. \tag{16}$$

For the map-synthesized scan, define the cyclically extended prefix-sum array with $S[0] = 0$:

$$S[t] = \sum_{j=1}^{t} \hat{z}_j'(\boldsymbol{p}), t \in [1, 2N_s], \tag{17}$$

where $\{\hat{z}_j'(\boldsymbol{p})\}$ is duplicated once to form the cyclic extension. Then, the mean range of the synthesized scan under orientation $\theta_m$ can be computed in constant time, as shown in Fig. 3(c), as

$$\bar{z}_{syn}'(\boldsymbol{p},\theta_m)=\frac{1}{\sum_k|\boldsymbol{I}_k|}\sum_{k=1}^{K}(S[v_k+m]-S[u_k+m-1]).\quad(18)$$

Finally, the SMAD at orientation $\theta_m$ is defined as

$$SMAD(\boldsymbol{p},\theta_m)=|\bar{z}'-\bar{z}_{syn}'(\boldsymbol{p},\theta_m)|.\quad(19)$$

And the final SMAD score can be expressed as

$$SMAD(\boldsymbol{p})=\begin{cases}SMAD(\boldsymbol{p},\theta_0), & (\sum_k|\boldsymbol{I}_k|)/N_s\geq 0.9\\ \min\limits_{\theta_m\in\Theta} SMAD(\boldsymbol{p},\theta_m), & otherwise\end{cases}.\quad(20)$$

In implementation, occasional invalid measurements (e.g., out-of-range or missing returns) may occur within each section. For efficient prefix-sum computation, these beams are assigned a constant placeholder value (e.g., zero). Since this handling is applied consistently to both real and synthesized scans, it does not introduce systematic bias in the SMAD evaluation. Moreover, as the metric is dominated by valid beams within the FoV, it remains robust to sparse invalid measurements.

*Remark 5*: In [35], the CAER is defined as

$$CAER(\boldsymbol{p},\theta_m)=\frac{1}{\sum_k|\boldsymbol{I}_k|}\sum_{k=1}^{K}\sum_{i\in\boldsymbol{I}_k}|z_i'-\hat{z}_{i+m}'(\boldsymbol{p})|.\quad(21)$$

Because the absolute operator lies inside the summation, prefix-sum queries cannot be applied, resulting in $O(\sum_k|\boldsymbol{I}_k|)$ complexity. In contrast, SMAD only depends on mean distance differences. With one $O(N_s)$ preprocessing step for prefix sums, each section query is $O(1)$, achieves lower complexity for non-panoramic LiDAR scan without rotational invariance.

Beyond computational efficiency, SMAD exhibits a different sensitivity to pose perturbations compared with CAER. Since CAER accumulates absolute differences at the beam level, even small translational offsets may introduce consistent deviations across many beams, leading to a noticeable increase in alignment error. In contrast, SMAD evaluates only the mean distance within each valid section. As a result, local deviations induced by small pose perturbations are aggregated before comparison. When the scene structure within a section is consistent, these deviations may partially cancel out at the mean level, making SMAD less sensitive to slight translational misalignment. This property is desirable for coarse hypothesis evaluation, where the goal is to capture section-level structural consistency rather than enforce exact beam-wise alignment. It allows SMAD to retain plausible candidates under pose uncertainty while avoiding premature rejection. In the proposed framework, this stage is followed by a likelihood-based fine evaluation, forming a hierarchical strategy that ensures robustness in early screening and reliability in final estimation.

### *C. Translation-Affinity Scan-to-Map Alignment Metric TAM*

For accurate scan-to-map alignment evaluation under a given pose hypothesis in localization or relocalization tasks, likelihood-field-based metrics are commonly used [18], [22], [33]. These metrics are based on the distance from each LiDAR beam endpoint to the nearest occupied cell on the grid map. Likelihood field-based metrics have higher sensitivity to alignment inconsistencies and are more scalable, making them more suitable for customized accurate alignment evaluation. Thus, in the proposed framework, the likelihood-field-based metric is applied in three stages. It first searches for the optimal orientation for each positional hypothesis to generate pose hypotheses (corresponding to the objective of (7)). It then provides the ranking basis for pose hypotheses. Finally, it evaluates the alignment score of refined poses after registration (corresponding to the objective of (9)). (7) is implemented in a discrete form as follows:

$$\theta_j^*=\underset{\theta_m\in\Theta}{argmax}\,\mathcal{F}\big(\boldsymbol{\mathcal{Z}}',(\boldsymbol{p}_j,\theta_m)\,\big|\,\boldsymbol{\mathcal{M}}\big).\quad(22)$$

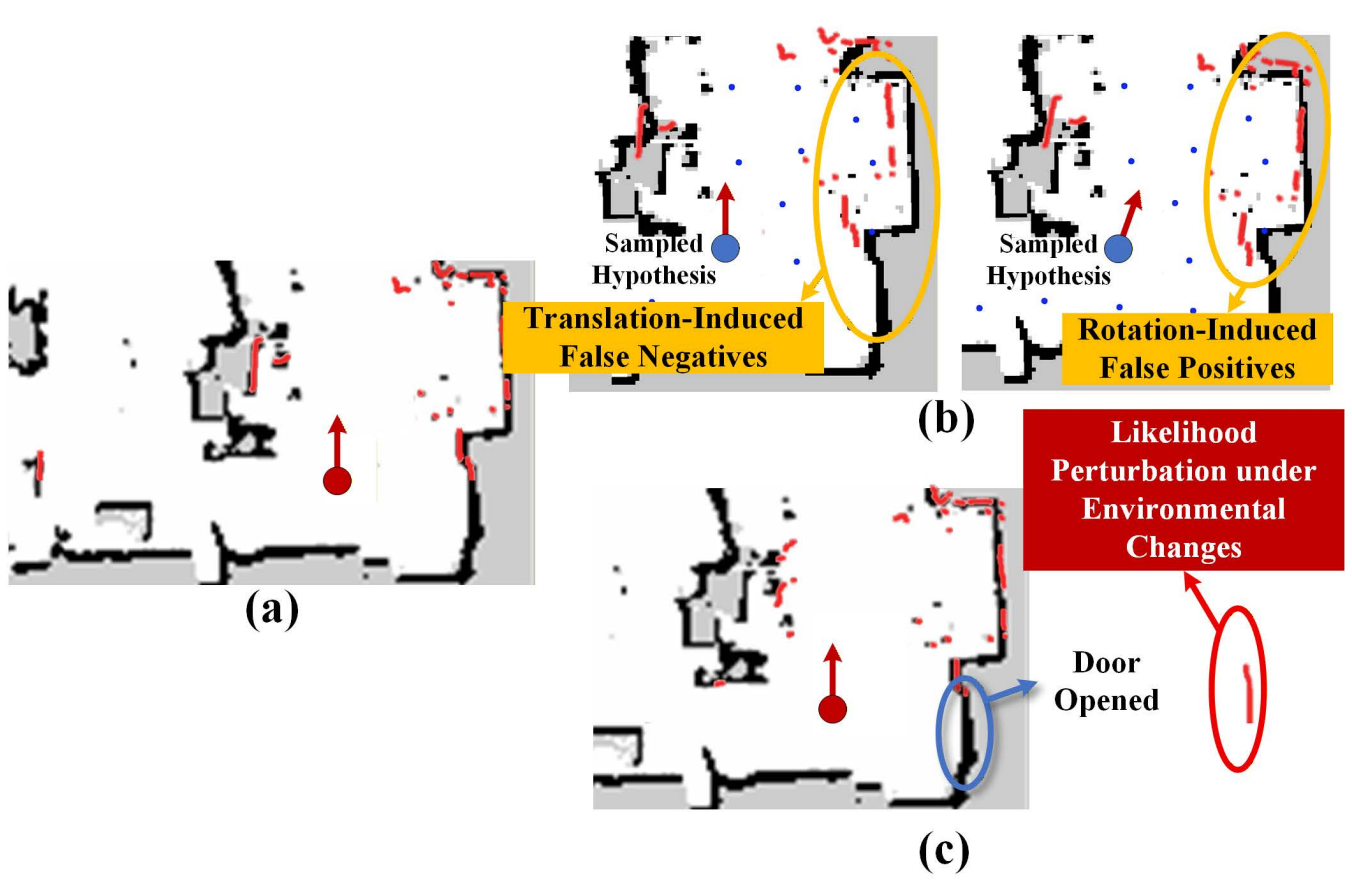

**Fig. 4.** Typical degradation cases of likelihood-field-based scan-to-map alignment metrics for relocalization tasks.

However, for orientation selection tasks under sparse positional hypothesis sampling and pose evaluation in complex real-world environments, the traditional likelihood-field-based metric [18], [33] tends to degrade. As shown in Fig. 4(a), under ideal conditions, the geometric correspondence between the scan and the map produces high-scoring alignment responses. When the positional hypothesis has a significant translational deviation $\Delta\boldsymbol{p}_j$ from the correct position, as shown in Fig. 4(b), the laser beams with the correct orientation become globally misaligned with structural boundaries and yield lower likelihood scores than incorrect orientations. This is translation-induced metric degradation. In real-world scenarios, even small-scale dynamic changes such as an opened door can cause local inconsistencies between the scan and the scene. As shown in Fig. 4(c), when the pose is correct, these local mismatches can sharply reduce the likelihood response. The local likelihood perturbation becomes more significant when the scan is non-panoramic.

To address the above issues, TAM is proposed. It possesses robustness in dynamic environments and maintains sensitivity to orientation estimation even in the presence of translational offsets, thereby exhibiting translation affinity. Let the hypothesized robot pose be denoted as $\boldsymbol{x}$. Each scan beam endpoint of $\boldsymbol{\mathcal{Z}}'$ in the map coordinate frame is represented by $\boldsymbol{q}_i(\boldsymbol{x})$. The likelihood-field model is defined as

$$\ell_i(\boldsymbol{x}) = log(z_{hit}exp(-\frac{d_{obs}^2(q_i(\boldsymbol{x}))}{2\sigma_{hit}^2}) + z_{rand} + \varepsilon), \quad (23)$$

where $z_{hit}$ is the weight of the hit component, which represents the probability that a beam endpoint coincides with the obstacle, $\sigma_{hit}$ controls the spatial decay of the likelihood field, $z_{rand}$ is the weight of the random measurement term, used to model unstructured noise, and $\varepsilon = 1e^{-6}$ is a small constant added for numerical stability. In dynamic environments, some beams yield abnormally low likelihood responses. Retaining these unstable beams can dominate the alignment statistics and excessively lower the matching score. To address this, TAM introduces a FoV-adaptive beam retention mechanism that balances information completeness and robustness across different FoV configurations. All beam responses are sorted in descending order:

$$\ell_1(\boldsymbol{x}) \geq \ell_2(\boldsymbol{x}) \geq \ldots \geq \ell_{N_Z'}(\boldsymbol{x}). \quad (24)$$

Let the ratio of the actual LiDAR FoV to the panoramic FoV $\lambda = \frac{\sum_k |\boldsymbol{I}_k|}{N_s}$, three retention ratios are defined as

$$\begin{cases} \alpha_{full} = 1 \\ \alpha_{mid} = \alpha_{mid}^{min} + \lambda(\alpha_{mid}^{max} - \alpha_{mid}^{min}), \\ \alpha_{low} = \alpha_{low}^{min} + \lambda(\alpha_{low}^{max} - \alpha_{low}^{min}) \end{cases} \quad (25)$$

with $\alpha_{full} > \alpha_{mid} > \alpha_{low}$. A larger FoV removes more low-likelihood beams to suppress unstable responses, while a smaller FoV retains more beams to preserve observation completeness. The corresponding beam numbers are

$$k_{full} = N_Z', k_{mid} = \lfloor \alpha_{mid} N_Z' \rfloor, k_{low} = \lfloor \alpha_{low} N_Z' \rfloor, \quad (26)$$

and the final robust response is obtained by

$$\bar{\mathcal{L}}(\boldsymbol{x}) = med\{\frac{1}{k_{full}}\sum_{i=1}^{k_{full}} \ell_i, \frac{1}{k_{mid}}\sum_{i=1}^{k_{mid}} \ell_i, \frac{1}{k_{low}}\sum_{i=1}^{k_{low}} \ell_i\}. \quad (27)$$

This adaptive filtering balances full, moderate, and strong exclusion modes. It enables TAM to automatically adjust to FoV variations and maintain stable alignment responses in dynamic or partially changed environments. Based on the robust likelihood response, the alignment likelihood term is defined as

$$\wp(\boldsymbol{x}) = exp(\bar{\mathcal{L}}(\boldsymbol{x})), \quad (28)$$

which represents the overall matching probability between the scan and the map structure. However, relying solely on the likelihood response makes stable normalization difficult. Because of the exponential decay of the likelihood-field model, even a few unmatched beams can cause a sharp decrease in the overall score. This makes the metric highly sensitive to local mismatches. To mitigate this effect, TAM introduces geometric distance statistics to complement the likelihood measure. After calculating the mean and variance of all distances $d_{obs}(\boldsymbol{q}_i(\boldsymbol{x}))$, $\bar{d}(\boldsymbol{x})$ and $\hat{\sigma}^2(\boldsymbol{x})$ are obtained. Here, $\bar{d}(\boldsymbol{x})$ reflects the average proximity of the scan to structural boundaries, and $\hat{\sigma}^2(\boldsymbol{x})$ reflects the consistency of distance distribution. To eliminate the influence of resolution and scale, the mean distance is clamped and normalized. Let the maximum valid distance be $d_{max}$ and the minimum effective distance be $d_{min} = r_{\mathcal{M}}/2$:

$$\tilde{d}(\boldsymbol{x}) = min(max(\bar{d}(\boldsymbol{x}), d_{min}), d_{max}),$$
$$s_d(\boldsymbol{x}) = 1 - \frac{\tilde{d}(\boldsymbol{x}) - d_{min}}{d_{max} - d_{min}} \in [0,1]. \quad (29)$$

The term $s_d(\boldsymbol{x})$ provides a smooth geometric correction to the likelihood response. It linearly maps the average distance into $[0,1]$, suppressing rapid likelihood decay caused by local mismatches and improving continuity and comparability across environments. However, both $\wp(\boldsymbol{x})$ and $s_d(\boldsymbol{x})$ remain strongly dependent on absolute distance magnitude and local occupancy proximity. Under sparse positional sampling or limited observability, this dependence may become unreliable. In particular, even when the orientation is incorrect, accidental local proximity to occupied structures may still produce a relatively high likelihood response or a small mean distance, resulting in overestimated scores. To address this issue, TAM introduces the structural consistency term:

$$c(\boldsymbol{x}) = exp(-\frac{1}{2}\hat{\sigma}^2(\boldsymbol{x})), \quad (30)$$

which measures the structural stability of distance distribution. When a translational offset exists but the orientation is correct, each term $d_{obs}(\boldsymbol{q}_i(\boldsymbol{x}))$ remains approximately equal to the translation magnitude. In this case, the variance $\hat{\sigma}^2(\boldsymbol{x})$ stays small, leading to an increase in $c(\boldsymbol{x})$. This compensates for the reduction of $\wp(\boldsymbol{x})$ and $s_d(\boldsymbol{x})$, thereby achieving translation affinity of the metric. Finally, the three components are fused through geometric averaging to obtain the normalized TAM:

$$\mathcal{T}(\boldsymbol{x}) = (\wp(\boldsymbol{x}) s_d(\boldsymbol{x}) c(\boldsymbol{x}))^{\frac{1}{3}} \in (0,1]. \quad (31)$$

*Remark 6*: In [33], the traditional likelihood-field-based scan-to-map alignment metric is simply defined as

$$\wp_{LF}(\boldsymbol{x}) = \sum_{i=1}^{N_Z'} \ell_i(\boldsymbol{x}). \quad (32)$$

The lack of normalization causes the metric values to depend strongly on the number of beams and the overall distance scale. This dependency prevents global comparability across distinct pose hypotheses. When sparse positional hypothesis sampling produces translational deviations, as shown in Fig. 4(b), the scan becomes globally misaligned with the map boundaries. Though the correct angle preserves geometric consistency, its mean distance to occupied cells increases, resulting in a lower likelihood score than that of incorrect orientations. This leads to translation-induced degradation of the metric. The proposed TAM mitigates these issues by introducing a structural consistency term and a normalization strategy based on the geometric mean.

*Remark 7*: The use of geometric averaging in (31) is motivated by both intuitive and mathematical considerations. The three components $\wp(\boldsymbol{x})$, $s_d(\boldsymbol{x})$, and $c(\boldsymbol{x})$ characterize complementary aspects of alignment quality, namely matching likelihood, average geometric proximity, and structural consistency. While $c(\boldsymbol{x})$ compensates for translation-induced bias at the component level, reliable pose evaluation still requires these properties to be jointly satisfied. Arithmetic averaging allows one component to compensate for the degradation of another, which may overestimate hypotheses with partial mismatch or structural inconsistency. In contrast, geometric averaging preserves multiplicative dependency among the normalized terms and suppresses the overall score when any component is low, thereby enforcing consistency across all aspects of the metric. This design aligns with the objective of TAM, where correct hypotheses are expected to exhibit simultaneously high likelihood, reasonable distance magnitude, and stable structural consistency. The efficacy of geometric averaging over arithmetic averaging is further validated by the ablation study in the experimental section.

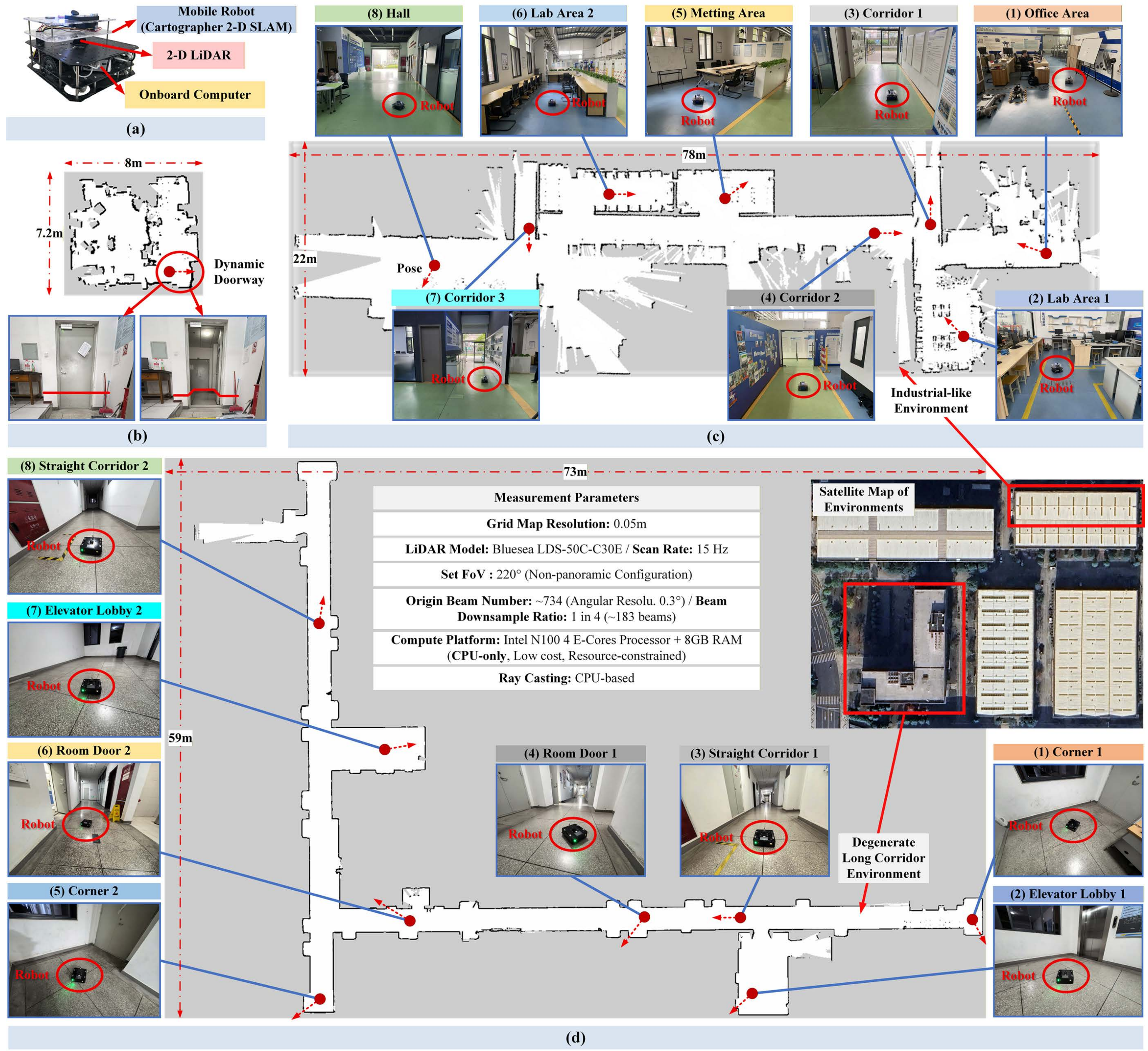

**Fig. 5.** Experimental robot and configurations of real-world experimental scenarios.

## IV. Experiments

### *A. Experimental Setup*

Fig. 5 shows the experimental setup and measurement parameters. The platform is a low-cost differential-drive robot equipped with an Intel N100 embedded processor (with only four E-cores) running Ubuntu 20.04, as shown in Fig. 5(a). A Bluesea LDS-50C-C30E LiDAR is used, operating at $15Hz$ with an effective range of $20m$. A non-panoramic FoV of 220° is adopted to reflect practical partial-observation conditions. The raw scan contains approximately 734 beams (0.3° angular resolution), which are downsampled to about 183 beams for efficiency. An occupancy grid map with a resolution of $0.05m$ is used, and scan synthesis is implemented via CPU-based ray casting. The robot radius is $r_{robot} = 0.21m$.

To evaluate the effectiveness of the proposed passive 2-D global relocalization framework, four sets of experiments were conducted, as described below. A demonstration video showcasing the relocalization performance of the proposed framework is available at https://youtu.be/fi7O4uxPjns.

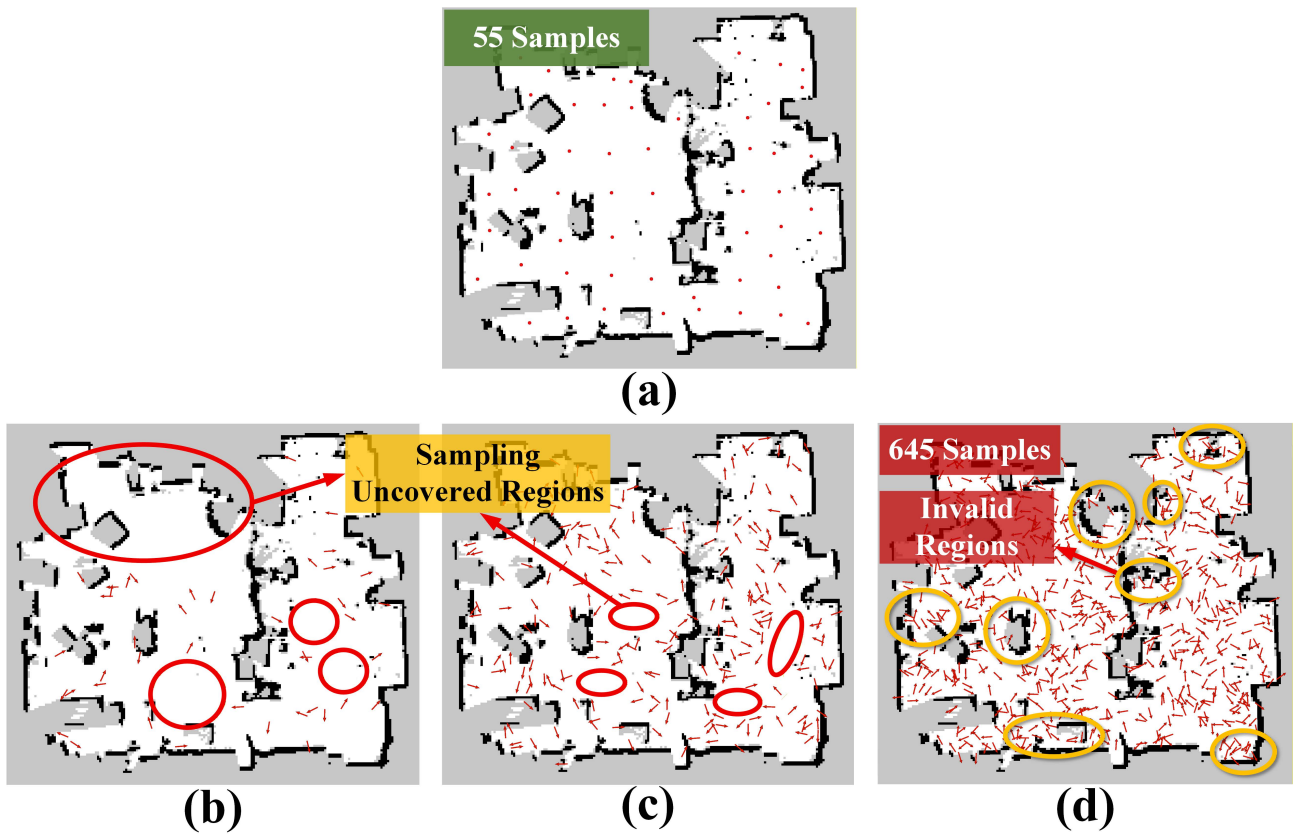

**Fig. 6.** Comparison of sampling methods for positional hypothesis generation under different hypothesis densities.

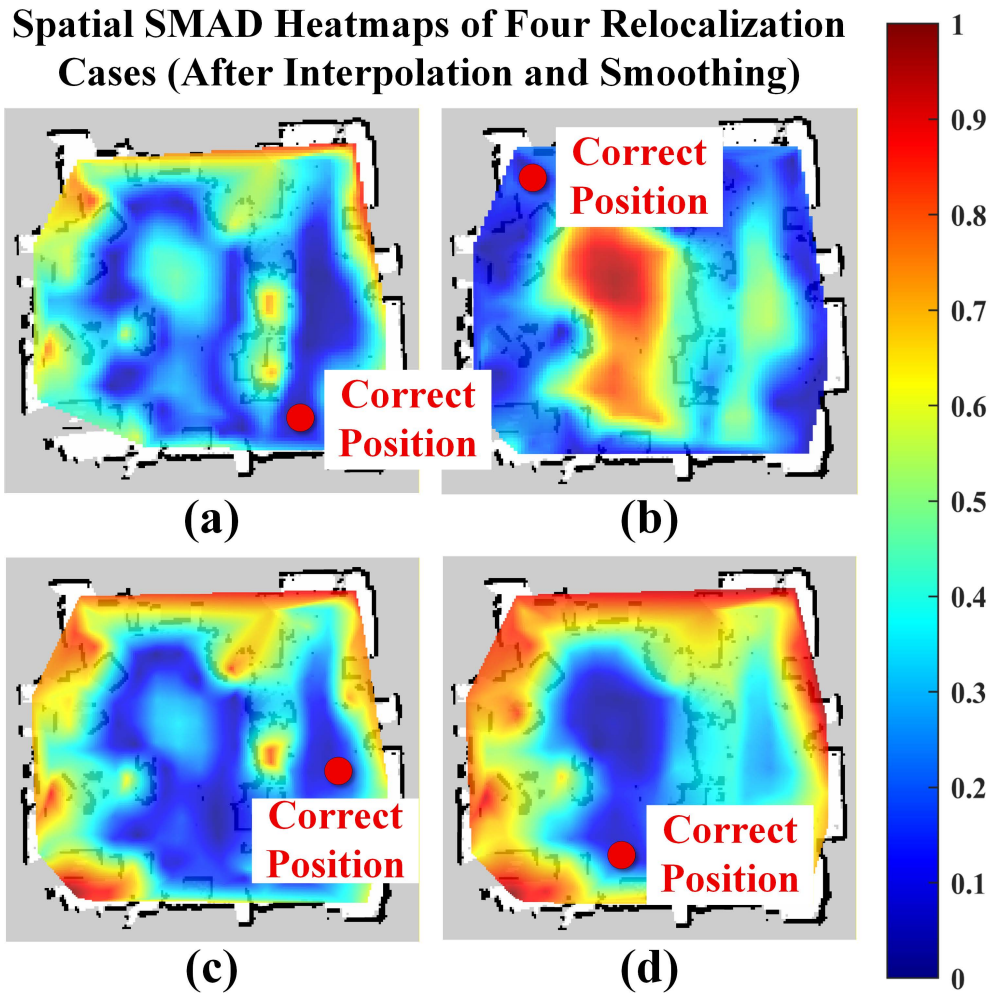

**Fig. 7.** Spatial SMAD Heatmaps of four relocalization cases in the scenario of Fig. 5(b).

*Case 1*: Functional validation of core parts. The small and cluttered environment in Fig. 5(b) was used to validate the core parts of the framework. Specifically, the experiments validated the sparse full-coverage hypothesis sampling capability of the traversability-constrained RRT, the inclusion of the SMAD of the correct position, and the discriminative capability of TAM under translational uncertainty and environmental changes.

*Case 2*: In the real indoor environments shown in Fig. 5(c) and Fig. 5(d), the proposed framework was compared with three advanced baselines: GMCL [19], FastGlobalLocalizer of Cartographer [23], which is based on BnB, and CBGL [35]. The two environments correspond to two different buildings, as indicated by the satellite views in Fig. 5. Rather than toy-level small scenarios, these environments correspond to full-floor layouts and are representative of the operational scale encountered in practical indoor mobile robot deployments.

Fig. 5(c) represents a large-scale industrial-like indoor environment with diverse functional areas, including offices, laboratories, meeting rooms, halls, and corridors. This environment is used to evaluate the overall applicability of the relocalization pipeline in representative indoor scenarios with varied geometric structures. In contrast, Fig. 5(d) corresponds to a different building dominated by long corridors, corners, room entrances, and elevator lobbies, and is characterized by a more structurally degenerate layout. This environment is used to evaluate the robustness of the compared methods in weakly distinguishable and corridor-dominated indoor scenarios.

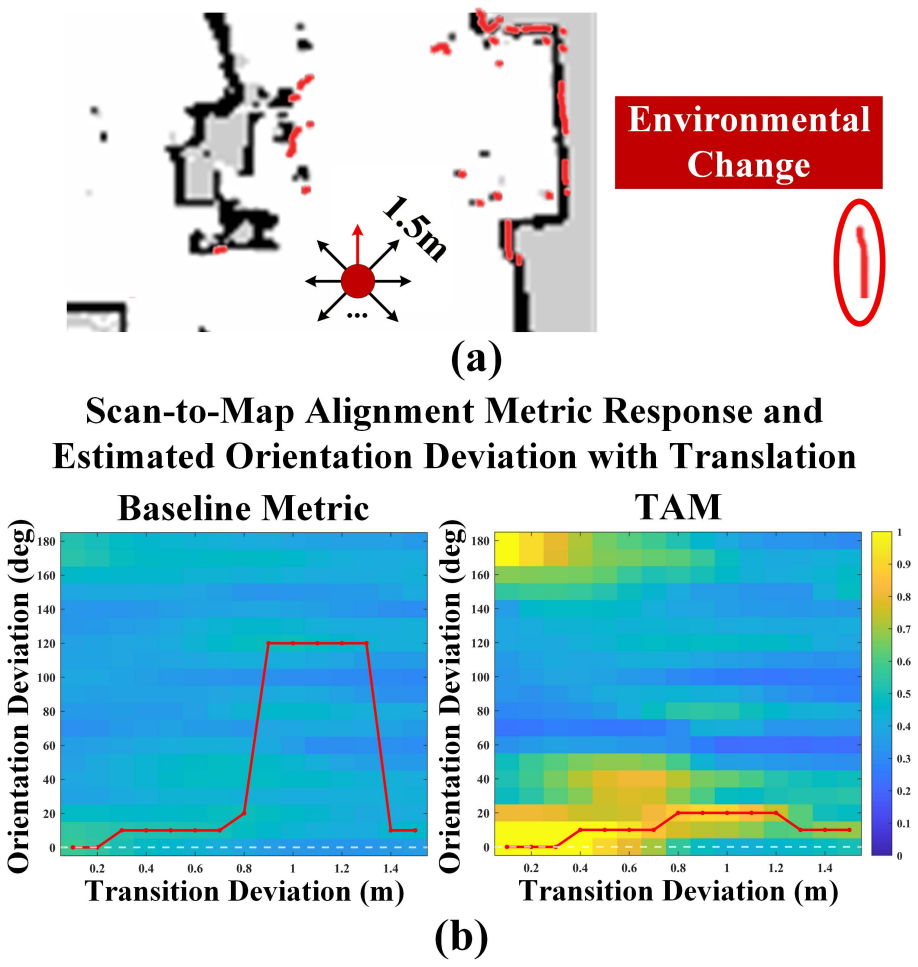

**Fig. 8.** Scan-to-Map Alignment Metric Response and Estimated Orientation Deviation with Translation (Normalized Likelihood vs TAM)

For each environment, eight representative poses were first recorded under valid SLAM tracking (using Cartographer 2-D) and physically marked on the floor. After initialization at the map origin, the robot was "kidnapped" to these poses in random order, and time-limited passive relocalization was performed using all baselines and the proposed method. The estimated poses were compared against the SLAM-tracked ground truth. All methods were evaluated in terms of success rate (SR), computational efficiency, and pose accuracy.

*Case 3*: To further analyze the contribution of each component, ablation studies were conducted in both the industrial-like environment in Fig. 5(c) and the degenerate long corridor environment in Fig. 5(d), using the same relocalization procedure as in Case 2.

First, an ablation study was performed on SMAD. Three configurations were evaluated: 1) without SMAD ordering, 2) with SMAD ordering but without prefix-sum acceleration, and 3) with both SMAD ordering and prefix-sum acceleration. Since the batched processing mechanism makes the overall success rates of these configurations similar in most cases, the average runtime, runtime standard deviation, and runtime interquartile range across all poses and repetitions were computed to evaluate the efficiency gain and runtime stability brought by SMAD ordering and prefix-sum acceleration. These reductions directly improve the ability of the overall pipeline to satisfy the fixed 30$s$ relocalization budget in challenging poses, while preserving similar success rates under the batched relocalization mechanism.

Second, an ablation study was conducted on TAM. The baseline metric is defined as

$$\wp_{baseline}(\boldsymbol{x}) = \left( exp\left(\frac{\wp_{LF}(\boldsymbol{x})}{N'_Z}\right) s_d(\boldsymbol{x}) \right)^{\frac{1}{2}} \in (0,1], \qquad (33)$$

which represents a normalized likelihood-field metric without the FoV-adaptive beam retention mechanism or the structural consistency term. In addition, to analyze the reasonability of the geometric fusion in TAM, an arithmetic-mean variant $TAM_{mean}$ is introduced:

$$\mathcal{T}_{mean}(\boldsymbol{x}) = \frac{\wp(\boldsymbol{x}) + s_d(\boldsymbol{x}) + c(\boldsymbol{x})}{3} \in (0,1]. \tag{34}$$

TABLE I
EVALUATIONS OF RELOC. METHODS IN REPRESENTATIVE ENVIRONMENTS (30S TIME BUDGET, INTEL N100, 220° FOV LIDAR)

| Pose | GMCL-8000 | | | | Cartographer-FastGlobalLocalizer | | | | CBGL- $\mathcal{D}$15 | | | | Proposed- $\mathcal{D}$1 | | | |
|---|---|---|---|---|---|---|---|---|---|---|---|---|---|---|---|---|
| | SR (%) | $t_{avg}$ (s) | $d_{avg}^{err}$ (m) | $\varphi_{avg}^{err}$ (°) | SR (%) | $t_{avg}$ (s) | $d_{avg}^{err}$ (m) | $\varphi_{avg}^{err}$ (°) | SR (%) | $t_{avg}$ (s) | $d_{avg}^{err}$ (m) | $\varphi_{avg}^{err}$ (°) | SR (%) | $t_{avg}$ (s) | $d_{avg}^{err}$ (m) | $\varphi_{avg}^{err}$ (°) |
| **Industrial-like Environment** | | | | | | | | | | | | | | | | |
| 1 | 50 | 16.891 | 0.077 | 3.713 | 100 | 16.821 | 0.091 | 0.426 | 90 | 3.156 | 0.051 | 1.883 | **95** | **0.466 (↓)** | **0.061** | **1.577** |
| 2 | 35 | 28.817 | 0.058 | 2.684 | 0 | / | / | / | 10 | 3.192 | 0.085 | 2.819 | **85 (↑)** | **1.597 (↓)** | **0.057** | **2.734** |
| 3 | 0 | / | / | / | 55 | 27.139 | 0.084 | 0.878 | 40 | 2.853 | 0.076 | 3.001 | **80 (↑)** | **0.992 (↓)** | **0.041** | **1.216** |
| 4 | 0 | / | / | / | 0 | / | / | / | 0 | / | / | / | **10 (↑)** | **1.533** | **0.045** | **1.347** |
| 5 | 10 | 18.648 | 0.063 | 3.176 | 100 | 20.068 | 0.079 | 1.001 | 90 | 3.101 | 0.069 | 2.245 | **100** | **0.901 (↓)** | **0.058** | **0.958** |
| 6 | 30 | 24.101 | 0.081 | 2.985 | 0 | / | / | / | 95 | 2.899 | 0.078 | 2.673 | **95** | **0.899 (↓)** | **0.096** | **0.841** |
| 7 | 0 | / | / | / | 90 | 29.336 | 0.036 | 0.593 | 35 | 2.596 | 0.093 | 2.118 | **100 (↑)** | **0.726 (↓)** | **0.089** | **2.493** |
| 8 | 5 | 27.005 | 0.094 | 2.012 | 100 | 11.393 | 0.023 | 0.227 | 0 | / | / | / | **100** | **0.613 (↓)** | **0.023** | **1.002** |
| **Degenerate Long Corridor Environment** | | | | | | | | | | | | | | | | |
| 1 | 0 | / | / | / | 100 | 9.756 | 0.028 | 0.647 | 0 | / | / | / | **100** | **0.334 (↓)** | **0.075** | **0.149** |
| 2 | 5 | 21.748 | 0.271 | 2.483 | 100 | 14.584 | 0.042 | 0.816 | 5 | 1.935 | 0.091 | 1.627 | **100** | **0.461 (↓)** | **0.058** | **1.757** |
| 3 | 0 | / | / | / | 0 | / | / | | 0 | / | / | / | **5 (↑)** | **0.873 (↓)** | **0.101** | **1.842** |
| 4 | 5 | 24.274 | 0.189 | 6.381 | 80 | 19.485 | 0.027 | 4.121 | 0 | / | / | / | **100 (↑)** | **0.374 (↓)** | **0.106** | **0.044** |
| 5 | 0 | / | / | / | 100 | 12.717 | 0.074 | 0.454 | 0 | / | / | / | **90** | **0.353 (↓)** | **0.016** | **0.831** |
| 6 | 15 | 25.382 | 0.225 | 5.246 | 0 | / | / | | 65 | 1.792 | 0.232 | 1.118 | **100 (↑)** | **0.437 (↓)** | **0.083** | **0.939** |
| 7 | 20 | 23.851 | 0.178 | 4.647 | 100 | 18.688 | 0.012 | 0.949 | 90 | 1.894 | 0.067 | 0.793 | **100** | **0.399 (↓)** | **0.071** | **0.198** |
| 8 | 0 | / | / | / | 0 | 9.756 | 0.028 | 0.647 | 90 | 1.889 | 0.466 | 0.562 | **95 (↑)** | **0.438 (↓)** | **0.041** | **0.833** |

Using $\wp_{baseline}$, $TAM_{mean}$, and $TAM$, the same relocalization procedure as in Case 2 was executed. The average success rate, average runtime, runtime standard deviation, and runtime interquartile range across all poses and repetitions were computed to evaluate the contribution of TAM and the rationality of the geometric-mean fusion strategy.

Finally, an ablation study was performed on the early-termination strategy. Both the presence or absence of early termination and multiple threshold settings were evaluated in the two environments. The average success rate, average runtime, runtime standard deviation, and runtime interquartile range were used to analyze the trade-off between relocalization reliability and computational efficiency under different termination conditions.

*Case 4*: To evaluate robustness under local environmental changes, an additional experiment was conducted in the dynamic doorway scenario shown in Fig. 5(b). The test pose is located near a doorway where the local map structure changes significantly between the door-closed and door-open conditions. Unlike Case 2, which focuses on floor-scale representative and degenerate indoor environments, this case is designed to isolate the effect of localized structural variation caused by a common indoor dynamic factor.

A reference map was first constructed under the door-closed condition. Passive relocalization was then performed at the marked pose under both door-closed and door-open conditions, as illustrated in Fig. 5(b). Following the same procedure as in Case 2, GMCL [19], FastGlobalLocalizer of Cartographer [23], CBGL [35], and the proposed method were evaluated by comparing the estimated poses against the SLAM-tracked ground truth.

*B. Functional Validation of Core Parts (Case 1)*

Fig. 6 compares the proposed traversability-constrained RRT sampling with random particle sampling for generating positional hypotheses. With a sampling spacing of $\rho = 1.0m$, corresponding to an expected hypothesis density of $\mathcal{D} = 1m^{-2}$, the proposed method achieves uniform and sparse coverage within the traversable region $\mathcal{R}(\boldsymbol{p}_0)$, as shown in Fig. 6(a), with $\varepsilon_{gain} = 2$ and $\eta_{max} = 0.6m$ . For random sampling, varying $\mathcal{D}$ controls the total number of samples. At $\mathcal{D} = 1$ and $\mathcal{D} = 5$, many uncovered areas remain, as shown in Fig. 6(b) and Fig. 6(c). When $\mathcal{D} = 15$, the coverage improves, as shown in Fig. 6(d), but the number of samples grows excessively, and many lie outside $\mathcal{R}(\boldsymbol{p}_0)$ , reducing subsequent search efficiency.

Fig. 7 shows spatial SMAD distributions for four different robot positions during relocalization. $n_{skip}^{beam} = 4$ . Regions with low SMAD values, indicating high scan-map consistency, are not unique, demonstrating the coarse nature of beam-error level metrics. However, the correct positions lie within low-SMAD regions. This demonstrates that after SMAD ranking, hypotheses near the correct positions are prioritized in early batches, improving overall relocalization efficiency.

Fig. 8 shows likelihood-field based scan-to-map metrics under translation and environmental changes. The scenario in Fig. 8(a) includes the door-opening variation of Fig. 5(b). Using 10° angular and $0.1m$ translational steps, $540$ test poses were generated from the correct pose in Fig. 8(a). $z_{hit} = 1.0, \sigma_{hit} = 0.2, and\ z_{rand} = 0$. Both the $\wp_{baseline}$ and TAM were computed for all hypotheses. As shown in Fig. 8(b), the heatmaps illustrate metric values, and the red curves indicate the orientation deviation of the maximum response for each

TABLE II

ABLATION STUDY ON SMAD

| $\rho$ | SMAD Enabled | Prefix-Sum Acceleration | $t_{avg}$ (s) | $t_{std}$ (s) | $t_{IQR}$ (s) |
|---|---|---|---|---|---|
| **Industrial-like Environment** | | | | | |
| 1.0 | No | No | 3.106 | 2.569 | 4.205 |
| | Yes | No | 1.271 | 0.381 | 0.612 |
| | **Yes** | **Yes** | **0.966 (↓)** | **0.372 (↓)** | **0.593 (↓)** |
| **Degenerate Long Corridor Environment** | | | | | |
| 1.0 | No | No | 1.573 | 1.478 | 2.071 |
| | Yes | No | 0.651 | 0.189 | 0.098 |
| | **Yes** | **Yes** | **0.459 (↓)** | **0.173 (↓)** | **0.086 (↓)** |

TABLE III

ABLATION STUDY ON TAM

| $\rho$ | Metric | $SR_{avg}$ (%) | $t_{avg}$ (s) | $t_{std}$ (s) | $t_{IQR}$ (s) |
|---|---|---|---|---|---|
| **Industrial-like Environment** | | | | | |
| 1.0 | $\wp_{baseline}(x)$ | 18 | 0.794 | 0.455 | 0.601 |
| | $TAM_{mean}$ | 78 | 0.954 | 0.425 | 0.564 |
| | $\boldsymbol{TAM}$ | **83 (↑)** | **0.966** | **0.407 (↓)** | **0.593** |
| 0.5 | $\wp_{baseline}(x)$ | 68 | 1.605 | 0.662 | 0.928 |
| | $TAM_{mean}$ | 83 | 1.694 | 0.698 | 1.017 |
| | $\boldsymbol{TAM}$ | **84 (↑)** | **1.723** | **0.709** | **0.996** |
| **Degenerate Long Corridor Environment** | | | | | |
| 1.0 | $\wp_{baseline}(x)$ | 24 | 0.331 | 0.162 | 0.123 |
| | $TAM_{mean}$ | 79 | 0.442 | 0.174 | 0.088 |
| | $\boldsymbol{TAM}$ | **86 (↑)** | **0.459** | **0.173 (↓)** | **0.086 (↓)** |
| 0.5 | $\wp_{baseline}(x)$ | 55 | 0.696 | 0.252 | 0.177 |
| | $TAM_{mean}$ | 82 | 0.748 | 0.311 | 0.182 |
| | $\boldsymbol{TAM}$ | **86 (↑)** | **0.784** | **0.291** | **0.193** |

translation. TAM remains stable with smaller orientation errors under translation. In addition, $\wp_{baseline}$ degrades notably under environmental changes.

### *C. Comparative Evaluations of Global Relocalization (Case 2)*

Relocalization experiments were conducted at eight poses in each of the two environments shown in Fig. 5(c) and Fig. 5(d), respectively. Each method was executed twenty times per pose, with the visiting order randomized. A trial was regarded as successful if the positional error satisfied $d_{err} < 0.5m$ and the angular error satisfied $\varphi_{err} < 30°$, which is sufficient for subsequent SLAM initialization. To ensure bounded computation, attempts exceeding the $30s$ time budget were counted as failures.

To improve fairness, each baseline was configured following its standard usage with only limited and justified adjustments. For GMCL, the particle number was set to 8000 as an empirical upper bound to avoid prolonged particle-update stalls, while the remaining settings were kept at their recommended values. Global relocalization for GMCL was performed by repeatedly triggering in-place particle updates. For FastGlobalLocalizer, the default localization configuration of Cartographer was used to reflect typical deployment practice. For CBGL, $n_{skip}^{beam} = 4$, and the parameters were set following the authors' recommended configuration. For both CBGL and the proposed framework, the number of orientation samples in (22) was set to 32. Following the settings in Case 1, CBGL used $\mathcal{D} = 15$, while the proposed framework used $\mathcal{D} = 1$. For the proposed framework, each batch contained 200 hypotheses and the local $top-20$ were retained, with an early-termination threshold of $\tau = 0.95$. Other parameters were identical to Case 1.

As shown in Table I, the proposed framework achieves the most balanced performance across the two environments, maintaining high success rates at most poses while remaining efficient, especially in the more challenging degenerate long corridor environment.

The results in Table I show that the proposed framework remains reliable across most poses in both environments. Reduced success rates occur only at a few weakly distinguishable poses, which is consistent with the observability limits discussed in Remark 1. Even so, the framework can still exploit subtle structural differences in many visually similar but not strictly symmetric indoor scenes, making passive relocalization effective in most representative indoor cases.

### *D. Ablation Studies (Case 3)*

Tables II to IV summarize the ablation results in the industrial-like environment and the degenerate long corridor environment. Table II shows that SMAD ordering consistently reduces runtime in both environments, and prefix-sum acceleration brings a further reduction while also suppressing runtime variation. This confirms that the proposed coarse-stage design improves both efficiency and runtime stability.

Table III evaluates the contribution of TAM. Compared with $\wp_{baseline}$, both the arithmetic-mean variant and the proposed TAM improve relocalization success rates in the two environments, confirming the benefit of the FoV-adaptive beam retention mechanism and the structural consistency term. Notably, the geometric-mean fusion consistently outperforms the arithmetic-mean fusion, which supports the proposed fusion strategy. Although TAM increases computation cost relative to $\wp_{baseline}$, the gain in relocalization reliability is substantial, and its performance remains stable under sparser hypothesis sampling.

TABLE IV

ABLATION STUDY ON EARLY-TERMINATION PARAMETER

| $\tau$ | $SR_{avg}$ (%) | $t_{avg}$ (s) | $t_{std}$ (s) | $t_{IQR}$ (s) |
|---|---|---|---|---|
| **Industrial-like Environment** | | | | |
| 0.91 | 75 | 1.201 | 0.802 | 1.235 |
| 0.93 | 77 | 1.442 | 0.681 | 1.009 |
| **0.95** | **83** | **1.694** | **0.698** | **1.017** |
| 0.97 | 83 | 1.872 | 0.702 | 1.112 |
| 0.99 | 84 | 1.997 | 0.715 | 1.103 |
| None | 84 | 2.113 | 0.582 | 0.966 |
| **Degenerate Long Corridor Environment** | | | | |
| 0.91 | 78 | 0.452 | 0.176 | 0.086 |
| 0.93 | 81 | 0.456 | 0.175 | 0.087 |
| **0.95** | **86** | **0.459** | **0.173** | **0.086** |
| 0.97 | 86 | 0.496 | 0.176 | 0.186 |
| 0.99 | 85 | 0.555 | 0.179 | 0.174 |
| None | 83 | 0.572 | 0.141 | 0.162 |

TABLE V

EVALUATIONS OF RELOC. METHODS IN DYNAMIC SCENARIO

| Door | Method | SR (%) | $t_{avg}$ (s) | $d_{avg}^{err}$ (m) | $\varphi_{avg}^{err}$ (°) |
|---|---|---|---|---|---|
| Close | GMCL-8000 | 35 | 3.574 | 0.097 | 4.582 |
| | FastGlobalLocalizer | 100 | 0.275 | 0.026 | 0.488 |
| | CBGL-$\mathcal{D}$15 | 75 | 0.737 | 0.057 | 2.276 |
| | **Proposed-$\boldsymbol{\mathcal{D}}$1** | **100** | **0.087 (↓)** | **0.018** | **0.561** |
| Open | GMCL-8000 | 10 | 5.832 | 0.136 | 4.978 |
| | FastGlobalLocalizer | 95 | 0.562 | 0.039 | 1.841 |
| | CBGL-$\mathcal{D}$15 | 50 | 0.711 | 0.063 | 1.921 |
| | **Proposed-$\boldsymbol{\mathcal{D}}$1** | **100 (↑)** | **0.098 (↓)** | **0.019** | **1.759** |

Table IV analyzes the early-termination strategy and the threshold parameter $\tau$. Early termination consistently reduces runtime in both environments. When $\tau$ is too small, the process tends to terminate too aggressively and the success rate decreases. When $\tau$ is too large, the behavior approaches that without early termination and the runtime increases. In the degenerate environment, the configuration without early termination does not achieve the highest success rate, suggesting that exhaustive search is not always beneficial under strongly multimodal conditions. The selected $\tau$ is therefore determined experimentally as a balance between relocalization reliability, computational efficiency, and runtime stability.

### *E. Evaluation in Dynamic Scenario (Case 4)*

Table V reports the relocalization results in the dynamic doorway environment shown in Fig. 5(b). The proposed framework maintains stable performance under both door states, while some baseline methods show a reduction in success rate or an increase in runtime after the door is opened. This result suggests that the proposed framework can remain effective not only in large indoor layouts and structurally challenging environments, but also under moderate localized environmental changes.

## V. Conclusion

This paper presents a passive 2-D LiDAR-based global relocalization framework. By integrating traversability-constrained RRT-based hypothesis sampling, SMAD-based coarse ranking, TAM for orientation selection and precise scan-to-map evaluation, and batched multi-stage inference with early termination, the proposed framework achieves a favorable balance between estimation reliability and computational efficiency. Real-world experiments conducted in three complex challenging scenarios demonstrate improved relocalization success rate and reduced computation time compared with existing advanced approaches, validating the effectiveness of the proposed framework under practical measurement constraints. This improves the robustness and reliability of LiDAR-based global pose estimation for long-term operation of mobile robotic systems. For future work, we will focus on further enhancing robustness in repetitive and low-feature environments, where measurement ambiguity and limited observability remain challenging.

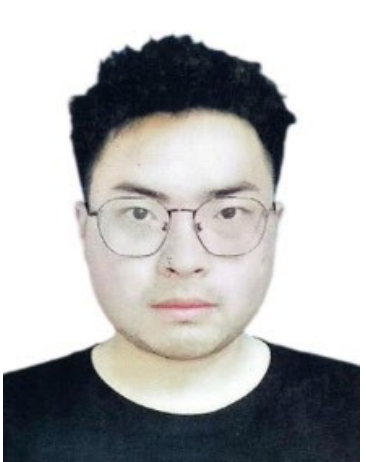

**Muhua Zhang** (Graduate Student Member, IEEE) received the bachelor's degree in mechanical design, manufacturing, and automation from Southwest Jiaotong University, Chengdu, China, in 2021, where he is currently pursuing the Ph.D. degree in control science and engineering.

His research interests include mobile robot perception, planning, and control, with a particular focus on mobile inspection robots operating in complex environments.

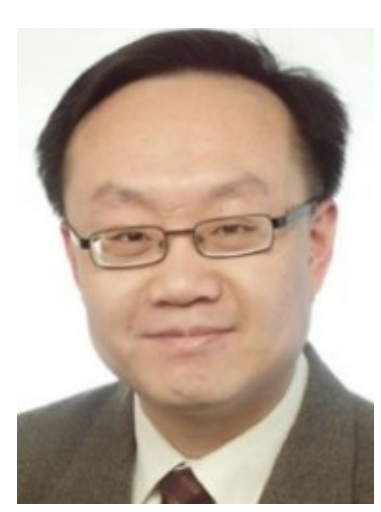

**Lei Ma** (Senior Member, IEEE) received the B.Eng. degree in automatic control from Chongqing University, Chongqing, China, in 1993, the M.Sc. degree in electrical engineering from Southwest Jiaotong University, Chengdu, China, in 1996, and the Dr.-Ing. degree in electrical engineering from Ruhr University Bochum, Bochum, Germany, in 2006.

From 2006 to 2009, he was an Assistant Professor with the Department of Robotics and Telematics, University of Würzburg, Würzburg, Germany, and also the Deputy Director and a Member of the Board of Supervisors with the Zentrum für Telematik e.V., Würzburg. Since September 2009, he has been a Professor at the School of Electrical Engineering, Southwest Jiaotong University. His research interests include control theory with applications to robotics, renewable energy, and rail transit systems.

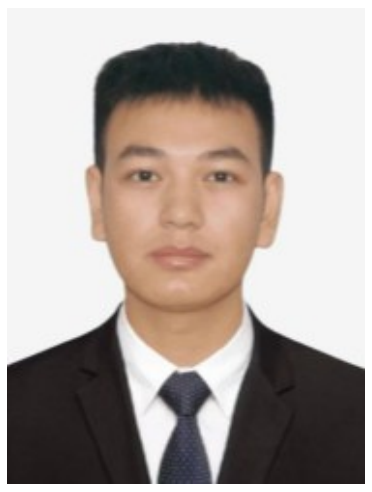

**Ying Wu** received the B.S. degree in measurement and control technology and instruments from Chongqing Technology and Business University, Chongqing, China, in 2019. He is currently pursuing the M.S. degree with the School of Electrical Engineering, Southwest Jiaotong University, Chengdu, China.

His research interests include robotic path planning and motion control.

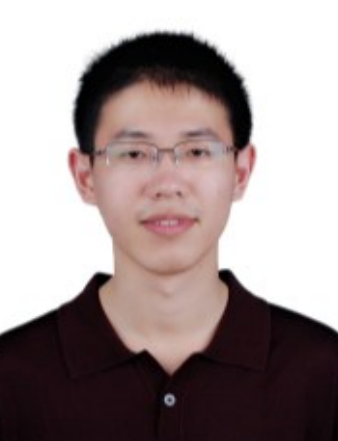

**Kai Shen** (Member, IEEE) received the B.S. and M.S. degrees from Northwestern Polytechnical University, Xi'an, China, in 2011 and 2014, respectively, and the Ph.D. degree in control science and engineering from Shanghai Jiao Tong University, Shanghai, China, in 2019.

From 2014 to 2015, he was an Assistant Engineer at Fujian Branch of China Telecom, Fuzhou, China. Before joining Southwest Jiaotong University, Chengdu, China, he was an Engineer at the Southwest Institute of Electronic Technology, Chengdu. He is currently an Associate Professor at the School of Electrical Engineering, Southwest Jiaotong University. His research interests include multitarget tracking and distributed fusion.

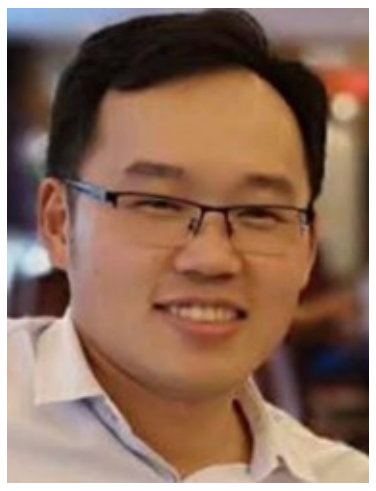

**Deqing Huang** (Senior Member, IEEE) received the B.S. and Ph.D. degrees in applied mathematics from the Mathematical College, Sichuan University, Chengdu, China, in 2002 and 2007, respectively, and the Ph.D. degree in control engineering from the National University of Singapore (NUS), Singapore, in 2011. From 2013 to 2016, he was a Research Associate with the Department of Aeronautics, Imperial College London, London, U.K. In 2016, he joined the Department of Electronic and Information Engineering, Southwest Jiaotong University, Chengdu, as a Professor and the Department Head.

His current research interests include modern control theory, data-driven control theory, and their applications.

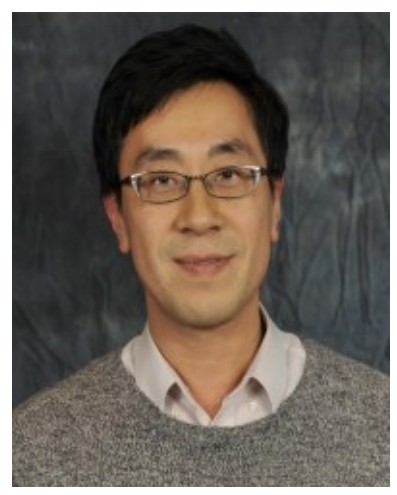

**Henry Leung** (Fellow, IEEE) received the B.Math. degree in applied mathematics from the University of Waterloo, Waterloo, ON, Canada, in 1984, the M.Sc. degree in mathematics from the University of Toronto, Toronto, ON, Canada, in 1985, and the M.Eng. and Ph.D. degrees in engineering physics and electrical engineering from McMaster University, Hamilton, ON, Canada, in 1986 and 1991, respectively.

Before joining the University of Calgary, Calgary, AB, Canada, he was a Defense Scientist with the Department of National Defence (DND), Ottawa, ON, Canada. He is currently a Professor with the Department of Electrical and Computer Engineering, University of Calgary. His research interests include information fusion, machine learning, the Internet of Things (IoT), nonlinear dynamics, robotics, and signal and image processing.

Dr. Leung is a Fellow of International Society for Optical Engineering (SPIE). He is the Topic Editor on Robotic Sensors of the *International Journal of Advanced Robotic Systems*. He is also the Editor of the book *Information Fusion and Data Science* (Springer).